# Evaluating Neural Cartographic Relief Shading for Urban Environments: A Downtown Calgary Study Using High-Resolution DEM and DSM Data

Emmanuel Stefanakis
Department of Geomatics Engineering, University of Calgary, Canada
emmanuel.stefanakis@ucalgary.ca

**Abstract**
This article explores the performance of analytical and neural-based hillshading methods in a dense urban environment using high-resolution digital elevation model (DEM) and digital surface model (DSM) data for downtown Calgary. The study compares single-direction and multi-direction analytical hillshading with relief shading generated in Eduard, a machine-learning system originally developed to emulate Swiss-style shaded relief trained primarily on mountainous landscapes. Because Eduard was not designed for buildings, bridges, streets, trees, and other urban infrastructures, the central question is not whether it perfectly reproduces urban morphology, but whether parameter tuning can nevertheless produce visually strong, cartographically useful, and in some cases superior results when compared with conventional analytical methods. The analysis focuses especially on terrain type, micro and macro generalization, and flat-area detail parameters, while keeping the large-scale shading style constant throughout the neural experiments. The article is structured as an exploratory comparison rather than a benchmark of universal best practice. It aims to identify where analytical hillshading remains more reliable, where Eduard offers unexpected strengths, and where neural shading fails because of its training bias toward alpine terrain. The study contributes to current work on terrain representation by testing whether a neural approach designed for natural landforms can be adapted to a highly built urban setting, and it concludes by arguing for future model training and evaluation specifically targeted at urban relief shading.

## 1. Introduction

Relief shading remains one of the most effective ways to communicate surface form in cartography because it transforms elevation data into a visually intuitive representation of depth, structure, and landform organization. Analytical hillshading methods are especially widespread because they are computationally efficient, transparent, and reproducible. Yet long-standing cartographic research has also shown that conventional analytical shading can be limited by directional bias, overly dark shadows, and a tendency to emphasize local slopes at the expense of larger landform structure (Imhof, 1982; Kennelly & Kimerling, 2004). These limitations become especially visible when relief shading is used not only for topographic terrain, but also for complex surface models that include abrupt and highly angular urban features.

Recent work in cartographic visualization has reopened the question of what constitutes high-quality shaded relief. Neural approaches such as Eduard extend earlier efforts to approximate the expressive qualities of expert manual shading by learning from examples rather than relying exclusively on analytical illumination models. The underlying research by Jenny et al. (2021) demonstrated that neural networks can reproduce important characteristics of manual relief shading, including local adaptation of illumination, suppression of unnecessary detail, and emphasis of larger terrain forms. However, these models were trained primarily on mountainous landscapes and Swiss-style relief (Hurni et al. 2025), which raises an important transferability problem when they are applied to urban environments dominated by buildings, engineered edges, transportation infrastructure, and relatively flat surfaces.

This study addresses that transferability problem using 20cm DEM and DSM data for downtown Calgary. Rather than assuming that Eduard would underperform in all urban contexts, the article investigates whether careful parameter tuning can produce useful and sometimes highly compelling urban shadings. The study compares single-direction and multi-direction analytical methods with multiple Eduard outputs while varying terrain type, micro and macro generalization, and flat-area detail settings. The large-scale shading style is kept constant to isolate the effects of those parameters. The objective is therefore twofold. First, to evaluate how these methods behave in an urban setting; and second, to identify both the strengths and the failure modes of Eduard in order to motivate future neural training specifically designed for urban morphology.

## 2. Data and Study Area

The study uses a pair of pixel-aligned, high-resolution elevation products covering a 9km² area in downtown Calgary (Figure 1), including a 20cm digital elevation model (DEM) and a resampled 20cm digital surface model (DSM). These elevation products were prepared for this study from a licensed 20cm resolution DEM created in 2024 and a 50cm resolution DSM created in 2022, both supplied by the City of Calgary (Reference codes: 2024 20cm DEM and 2022 50cm DSM for TWP 24 R01 W5 S15, 16, 21, and 22).

The DSM represents the top-most reflective surface and therefore includes buildings, bridges, vegetation, and other urban structures, while the DEM represents the bare-earth surface after removal of above-ground objects. Together, these two datasets make it possible to compare how hillshading methods respond to fundamentally different conceptions of urban surface form, one that foregrounds built morphology and one that isolates terrain.

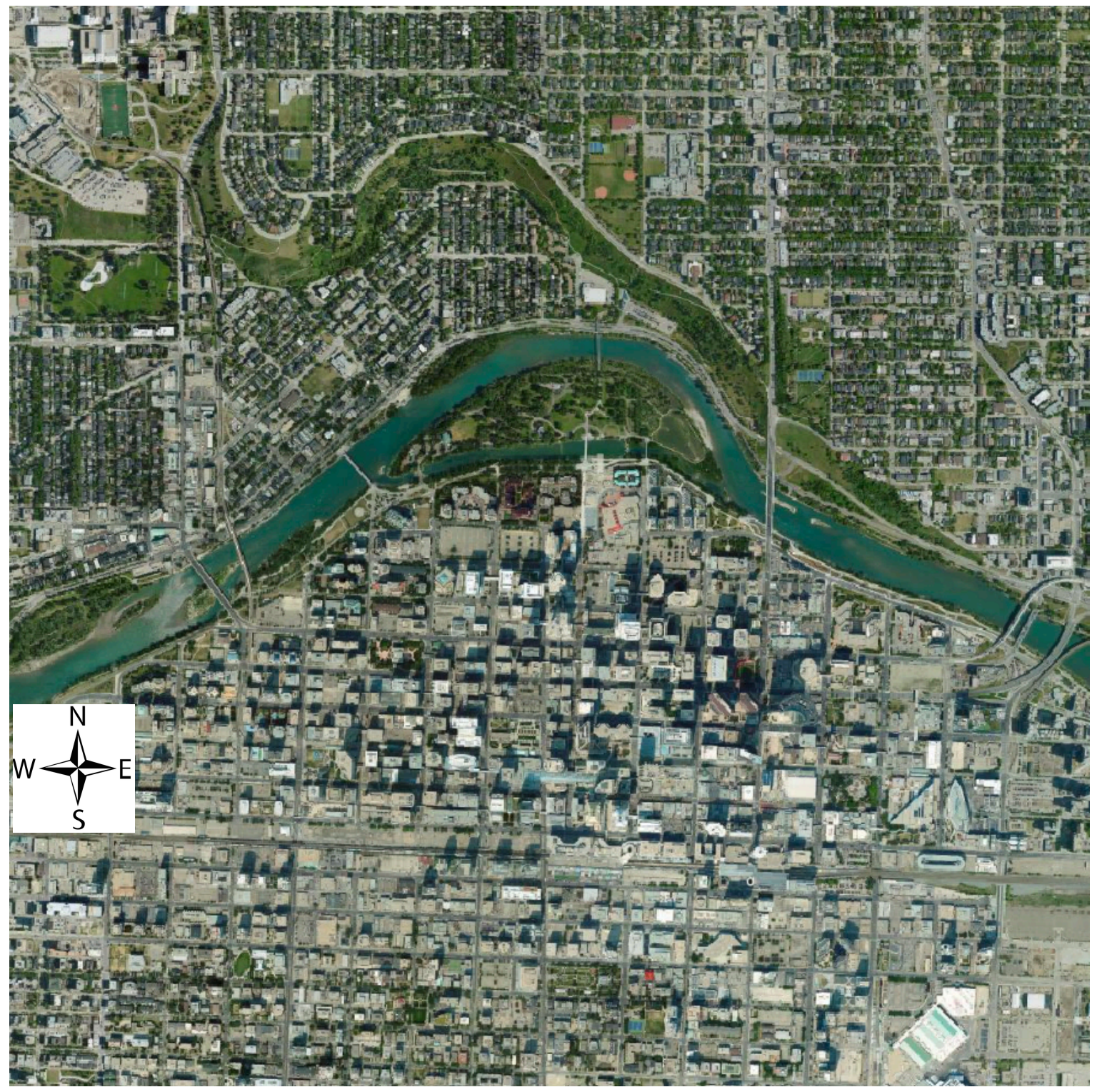

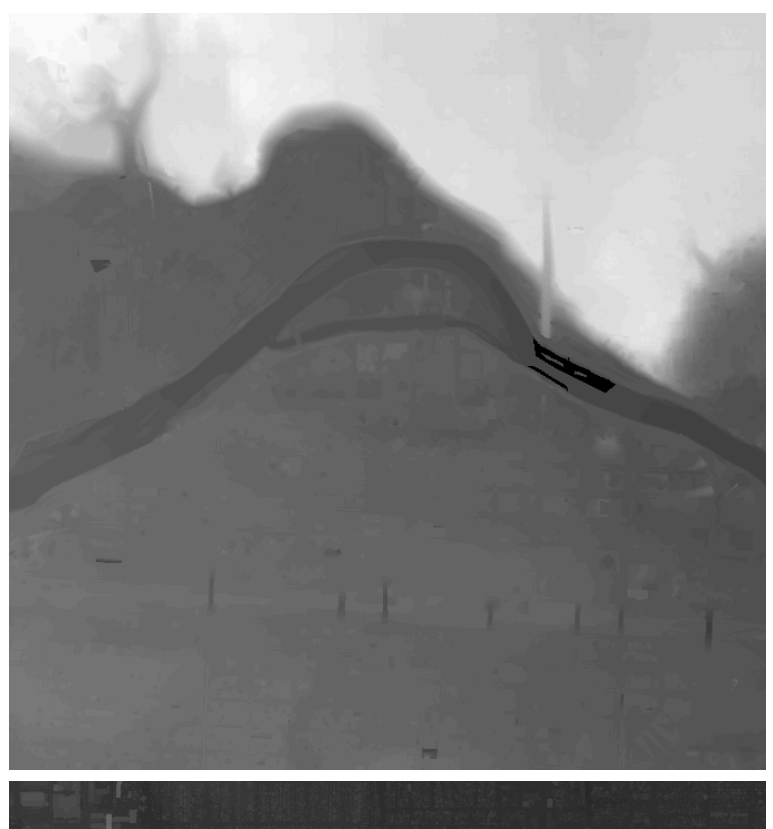
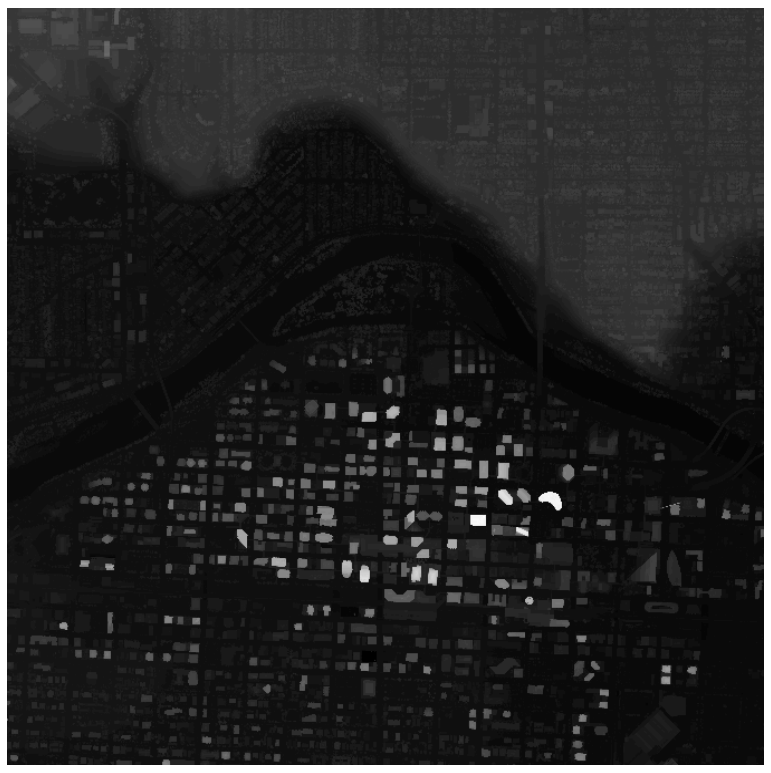

**Figure 1.** Left: the 3km x 3km downtown Calgary study area at scale 1:30,000 (Basemap imagery: Esri World Imagery). Top-right: the 20cm DEM of the same area at scale 1:60,000.; range of elevation values from 1020m (black) to 1091m (white). Bottom-right: the 20cm DSM of the same area at scale 1:60,000; range of elevation values from 1032m (black) to 1290m (white).

Downtown Calgary offers a particularly relevant test case because it combines dense high-rise development with streets, overpasses, open paved areas, small buildings and residential houses, river-adjacent terrain, trees, a well-defined valley and pronounced slopes along the river's north bank, as well as subtle ground variation south of the river that may be difficult to visualize when vertical differences are small. In such settings, the visual quality of hillshading depends not only on relief amplitude, but also on the cartographic handling of hard edges, flat surfaces, and abrupt transitions between natural and anthropogenic forms. The high spatial resolution of the available datasets can preserve fine urban detail, but it also increases the likelihood that shading will become noisy, overly literal, or visually cluttered if the representation is not generalized appropriately.

## 3. Methodology

The methodology is designed as a comparative cartographic experiment. This study does not attempt to evaluate hillshading solely through algorithmic accuracy metrics. Instead, it examines how well each method produces visually legible and cartographically convincing representations of urban form. In other words, the goal is to understand which outputs best communicate relief, structure, and urban morphology, and under what parameter settings they do so.

This study applied both analytical and neural-based cartographic relief shading methods (Burrough and McDonnell, 1998; Stefanakis, 2026a). The analytical hillshading experiments served as the baseline condition, with single-direction hillshading representing the standard analytical approach. Because Calgary is located in the Northern Hemisphere, a southwestern illumination direction, with a nominal light azimuth of 225°, was adopted instead of the traditional northwestern convention (Biland and Çoltekin, 2017). As discussed in Stefanakis (2026b), this choice produces more natural solar lighting and more realistic relief shading, making it easier for map readers to compare the results with satellite imagery from online map service providers such as Google and Esri (Figure 1; notice the shadows of the buildings caused by southern sunlight). Given the tall buildings in the downtown core, some exceeding 200m, and the small extent of the study area (3km × 3km), no vertical exaggeration was applied (that is, the z-factor was set to 1 in all experiments). Multi-direction hillshading was used as the enhanced analytical comparison because it reduces the directional bias of single-source illumination and often reveals landforms more evenly across surfaces with different orientations. This is achieved by setting the main/nominal light azimuth, e.g., 225°, while the rendering algorithm uses multiple directions internally, like 180°, 225°, 270°, and 315°. Together, the two analytical hillshading methods provide transparent, reproducible, and interpretable reference cases against which neural-based hillshadings implemented in Eduard (Eduard, 2025) are evaluated.

The Eduard experiments constitute the exploratory core of this study. In these experiments, the large-scale shading style was held constant because of the high resolution of both the DEM and DSM, while the terrain type, micro-generalization, macro-generalization, and flat-area detail parameters were varied systematically. This approach made it possible to assess whether a model trained on mountainous shaded relief could be adapted to suppress alpine exaggeration, reduce visual clutter, and recover useful detail in dense and diverse urban environments. A range of parameter combinations was tested to identify both the strengths and limitations of neural-based hillshading in urban scenes.

For the purposes of this study, a series of hillshades was generated for both the DEM and DSM using analytical and neural-based methods and testing a range of parameter settings. To evaluate the quality of the hillshades and determine the optimal parameter values for each elevation product, a cartographic comparison was undertaken. This assessment combined visual inspection, side-by-side interpretation, and explicit evaluation criteria, including landform clarity, readability of urban features, tonal balance, noise suppression, edge harshness, and overall cartographic suitability. An evaluation panel consisting of cartographers, GIS analysts, and graduate students was involved in the assessment. This qualitative review provides an initial interpretive framework for judging the practical usefulness of Eduard-generated hillshades in an urban mapping workflow. The following section summarizes the results of the assessment through a gallery of representative hillshade excerpts.

## 4. Results

This section presents the experimental results in a logical sequence, moving from baseline analytical methods to neural-based outputs. It first examines the analytical single-direction and multi-direction hillshades, before introducing the Eduard results generated from both the DEM and the DSM. Owing to space limitations, the discussion focuses on selected excerpts that most effectively demonstrate the influence of different Eduard parameter values. The analysis addresses both the study area as a whole and selected urban morphological forms that are uncommon in regional and mountain environments.

Figure 2 presents hillshades of the terrain for the entire study area generated using single-direction hillshading, multi-direction hillshading, and a neural-based hillshade produced by Eduard. All three hillshades were generated with an illumination azimuth of 225°, representing solar illumination from the southwest on the DEM. For the two analytical methods, the illumination altitude was set to 45°. For the neural-based hillshade, parameter values were selected to better represent both the major landforms and the subtle terrain details (Stefanakis, 2026c). Specifically, the Large shading style was chosen to suit the high-resolution DEM, together with a moderate terrain type value of 30, a generalization macro value of 50, and a flat-area detail value of 50. No aerial perspective was applied. Readers consulting the digital version may wish to enlarge the excerpts to better discern the differences.

Comparing the two analytical hillshadings, multi-directional hillshading reduces azimuth dependence, making the major landform details along the northern riverbank southwest-facing slopes more consistently visible instead of allowing them to disappear in the excessive brightness. However, compared with single-direction hillshading, this more balanced illumination does not seem to provide a better representation of the riverbank in its entirety. In contrast, neural-based shading provides a more effective representation of relief across the entire study area. Both major landforms and subtle details in flatter areas are clearly discernible, and the terrain patterns created by the street network are rendered efficiently as shown in Figure 3.

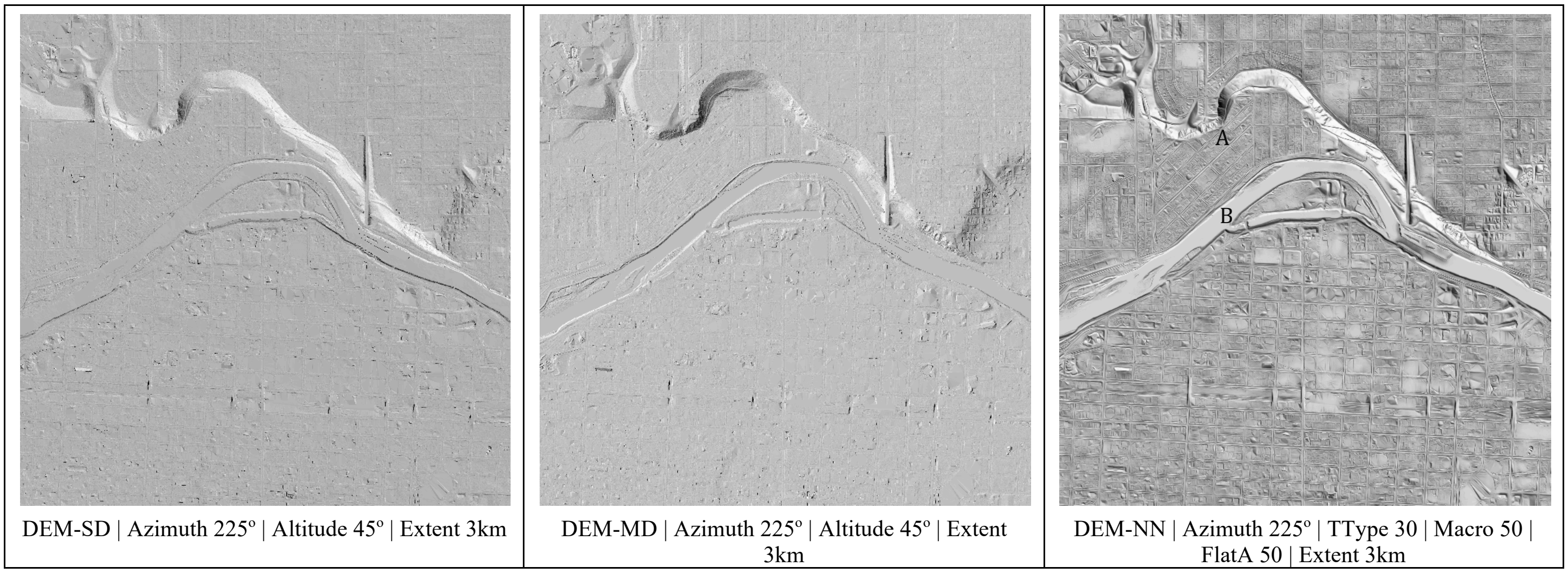

**Figure 2.** From left to right: Single-direction hillshade; Multi-direction hillshade; and Neural hillshade on the DEM for the entire study area at scale 1:50,000.

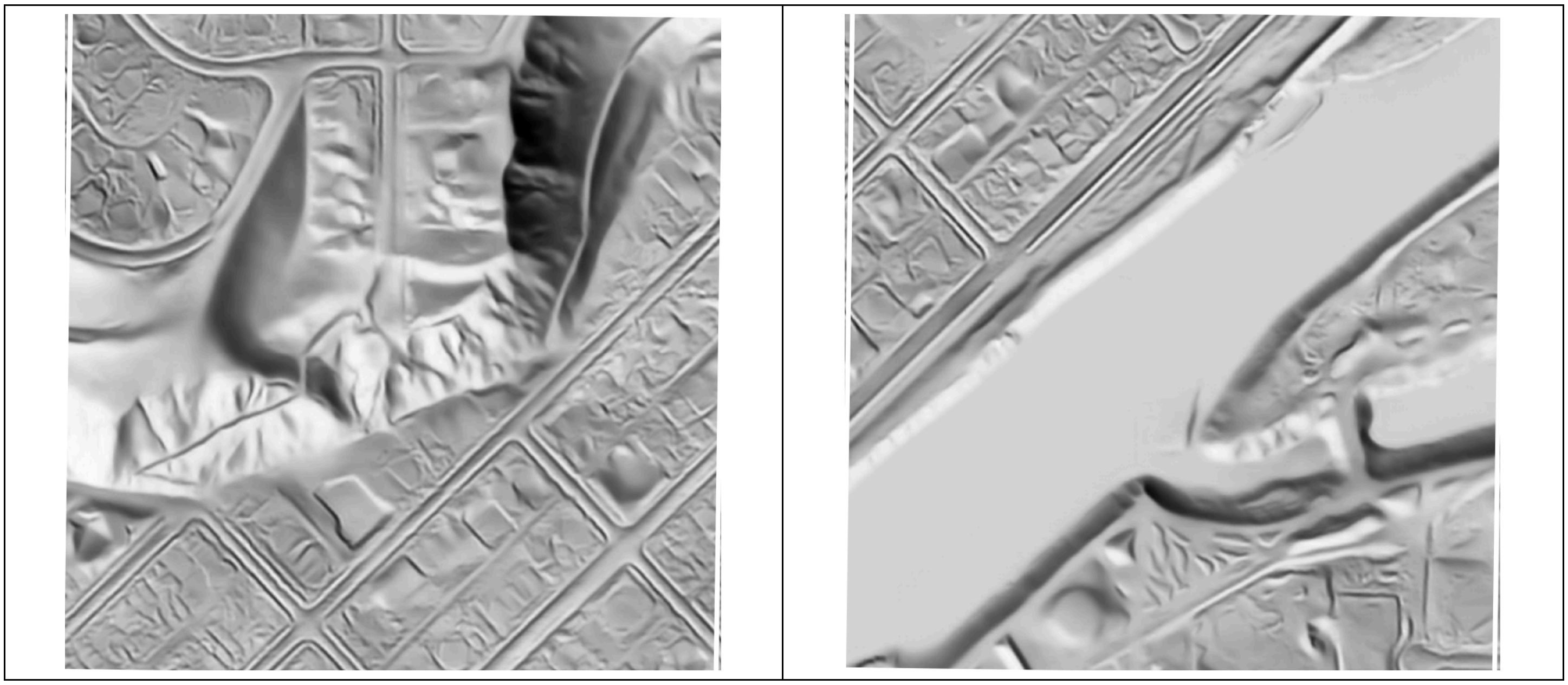

**Figure 3.** Excerpts from the neural hillshade of the study area near locations A and B in Figure 2 at scale 1:5,000.

Figure 4 shows hillshades of the DSM surface for the entire study area generated using single-direction hillshading, multi-direction hillshading, and a neural-based hillshade produced by Eduard. All three were generated with an illumination azimuth of 225°, representing solar illumination from the southwest on the DSM. For the analytical methods, the illumination altitude was set to 45°. For the neural-based hillshade, parameters were selected to highlight both major landforms and fine surface detail. The Large shading style was used to match the high-resolution DSM, with a terrain-type value of 60. No generalization, flat-area detail enhancement, or aerial perspective was applied. As discussed below, these settings are not equally suitable for all surface structures, and shading quality can be improved through parameter adjustment.

The following subsections and figures present how analytical and neural hillshading methods depict urban infrastructure, including both grey infrastructure (i.e., built systems such as roads, bridges, utilities, drainage, and transit networks) and green infrastructure (e.g., trees, parks, and other vegetation). As the relief shading generated with Eduard was not designed for buildings, bridges, streets, trees, or other forms of urban infrastructure, the key question is not whether it can perfectly reproduce urban morphology, but whether parameter tuning can nonetheless yield visually compelling, cartographically useful, and in some cases superior results compared with conventional analytical methods. The analysis focuses on terrain type, micro- and macro-generalization, and flat-area detail parameters, while keeping the large-scale shading style constant throughout the neural experiments.

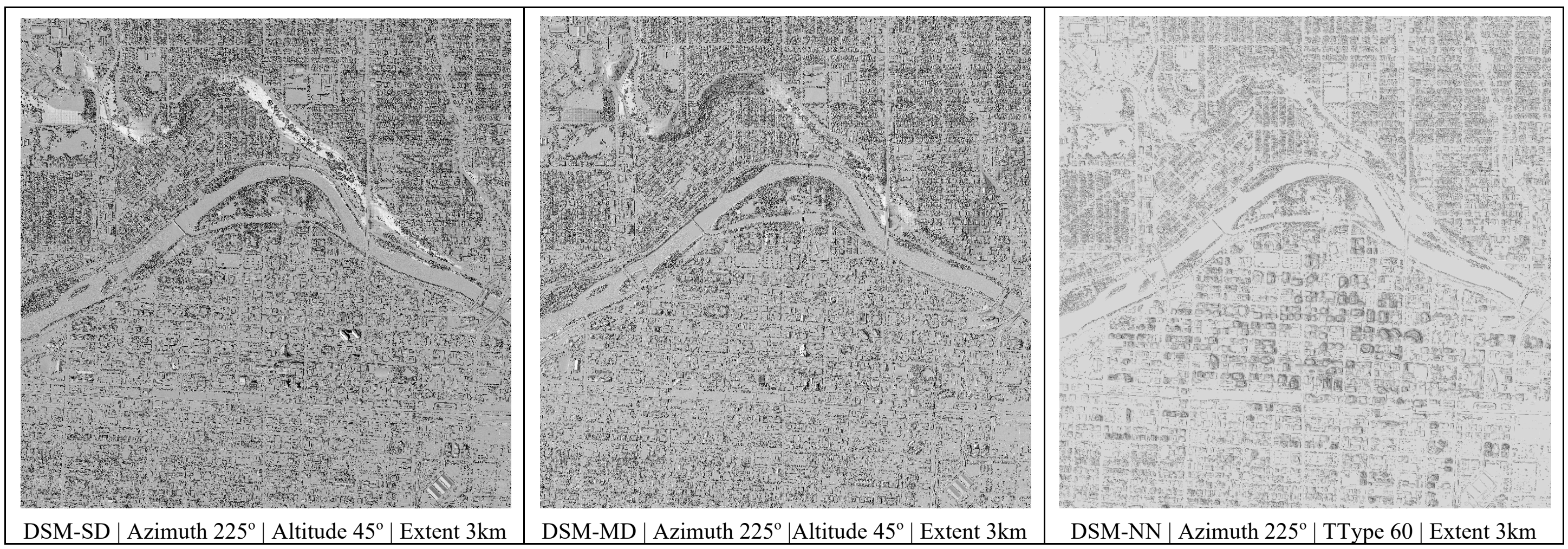

**Figure 4.** From left to right: Single-direction hillshade; Multi-direction hillshade; and Neural hillshade on the DSM for the entire study area at scale 1:50,000.

### 4.1 Shading green infrastructure

Figure 5 depicts excerpts from satellite imagery and hillshades of the McDougall Centre, an example of the Beaux-Arts style, and its gardens on 4 Avenue SW, at a scale of 1:2,000. A medium terrain type value of 60 produces smooth shading of the garden trees, but it does not clearly represent the architectural form of the gable roof, the porch, or the garden paths, all of which are more distinctly depicted in the analytical hillshades. This suggests that, while the neural-based hillshade can effectively generalize vegetation and produce visually coherent shading, it may suppress finer architectural and urban landscape details (also refer to Section 4.3) at this scale. In contrast, the analytical hillshades preserve sharper edges and stronger shadow contrasts, allowing the roof structure, building elements, and garden layout to remain more legible.

Satellite image | Esri World Imagery | Extent 180m x 100m

DSM-SD | Azimuth 225º | Altitude 45º | Extent 180m x 100m

DSM-MD | Azimuth 225º | Altitude 45º | Extent 180m x 100m

NN-DSM | Azimuth 225º | TType 60 | Extent 180m x 100m

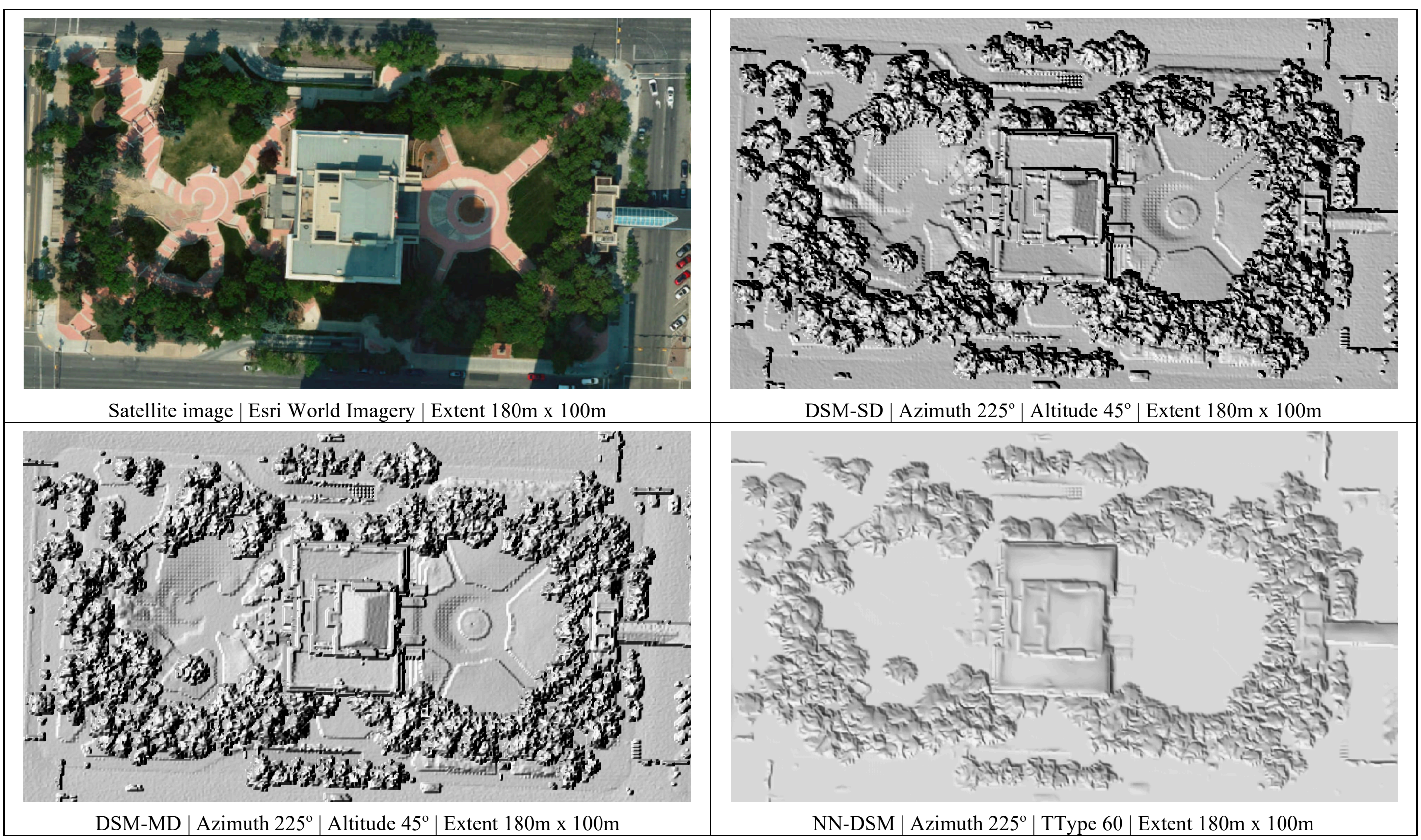

**Figure 5.** Excerpts of Satellite image; Single-direction hillshade; Multi-direction hillshade; and Neural hillshade on the DSM of McDougall Centre and gardens on 4 Ave SW at scale 1:2,000.

### 4.2 Shading bridges and surrounding infrastructure

Figure 6 presents excerpts from satellite imagery and hillshades of the Peace Bridge and the western bank of Prince's Island, at a scale of 1:3,750. Both analytical methods clearly depict the vaulted ceiling and the openings on either side of the bridge, the passage connecting the island to the south bank on the right side of the excerpts, and the pedestrian and cycling paths at the southern end of the bridge. Multi-directional shading provides a clearer representation of the steep riverbank at the northern end of the bridge, as it balances shadows more effectively. Tree distribution is evident in both analytical hillshades, although it appears coarse and is not represented in a particularly naturalistic manner.

The neural-based hillshades also provide a good overall representation of the bridge and the passage, especially when the terrain type is set to 90 and macro-generalization is set to 30. The trees appear less coarse and more natural, however, the paths and, in particular, the steepness of the northern riverbank are less visible.

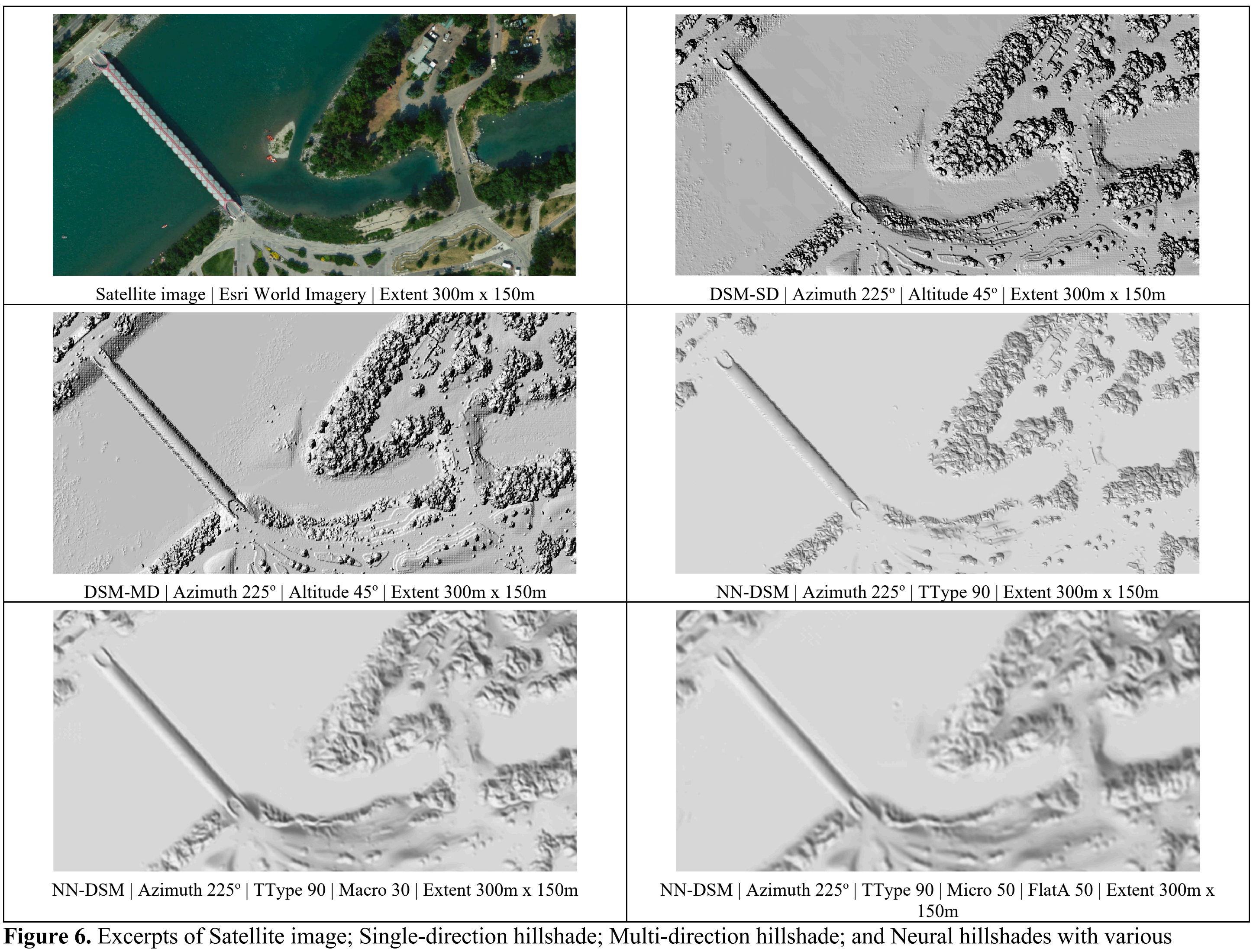

**Figure 6.** Excerpts of Satellite image; Single-direction hillshade; Multi-direction hillshade; and Neural hillshades with various parameter values on the DSM for the Peace Bridge and the western bank of Prince's Island at scale 1:3,750.

The Historical Reconciliation truss bridge and the adjacent flyover bridge are located on the eastern riverbank of the study area. Figure 7 presents excerpts from the satellite imagery and hillshades at a scale of 1:3,000. The analytical hillshades are highly effective in depicting the flyover bridge, as well as the buildings and street infrastructure on the southern bank. Neural-based hillshading, using a medium terrain-type value of 60 and a micro-generalization value of 30, also produces a strong overall shading result. In particular, the beams of the truss bridge are rendered very clearly.

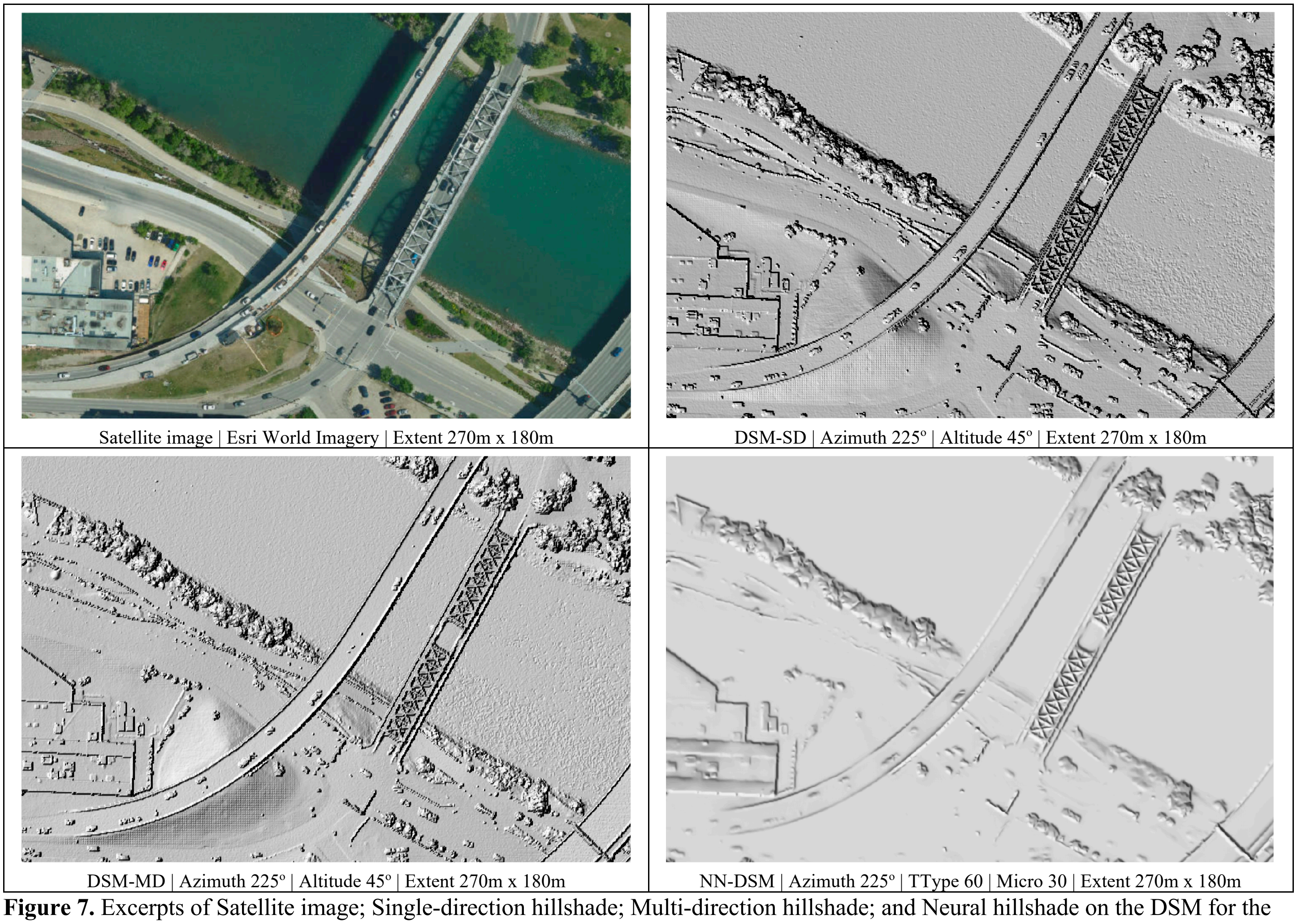

**Figure 7.** Excerpts of Satellite image; Single-direction hillshade; Multi-direction hillshade; and Neural hillshade on the DSM for the Historical Reconciliation (truss) bridge and the adjacent flyover bridge at the East of the downtown at scale 1:3,000.

### 4.3 Shading low-rise residential houses and buildings

The northeastern part of the study area is dominated by low-rise residential buildings, primarily with gable roofs, which consist of two sloping planes that meet along a central ridge and form a triangular wall at each end. A smaller number of buildings have hip roofs, comprising two trapezoidal roof planes and two triangular roof planes that meet along five ridges. Figure 8 presents an excerpt of two building blocks located north of 8 Ave NE and east of 4 Street NE at a scale of 1:2,000. Both analytical methods produce hillshades that clearly delineate the ridges of the gable and hip roofs. The neural-based hillshade with a high terrain type value of 90 smooths the ridges, producing a more continuous surface while still accurately representing the roof forms at this scale. It also provides more balanced shadowing across the hillshade. In addition, the trees appear more natural in the neural-based output.

Figure 9 shows excerpt satellite imagery and hillshades of the complex compound hip roof of Grace Presbyterian Church on 9 Street SW at a scale of 1:1,600. As with the buildings shown in Figure 8, the neural-based hillshade smooths the roof ridges and produces softened shadows. However, at this larger scale, the roof depiction loses some of its rigid architectural form, which is clearly visible in the analytical hillshades. Reducing the terrain type value to 30 makes the roof ridges and overall roof form less discernible, while applying micro-generalization does not appear to produce any noticeable improvement when the terrain type value is set to 90. The neural-based hillshade captures the church tower at the northeastern corner of the edifice and the surrounding infrastructure, although less effectively, while representing the surrounding trees more naturally.

Satellite image | Esri World Imagery | Extent 170m x 100m

DSM-SD | Azimuth 225º | Altitude 45º | Extent 170m x 100m

DSM-MD | Azimuth 225º | Altitude 45º | Extent 170m x 100m

NN-DSM | Azimuth 225º | TType 90 | Extent 170m x 100m

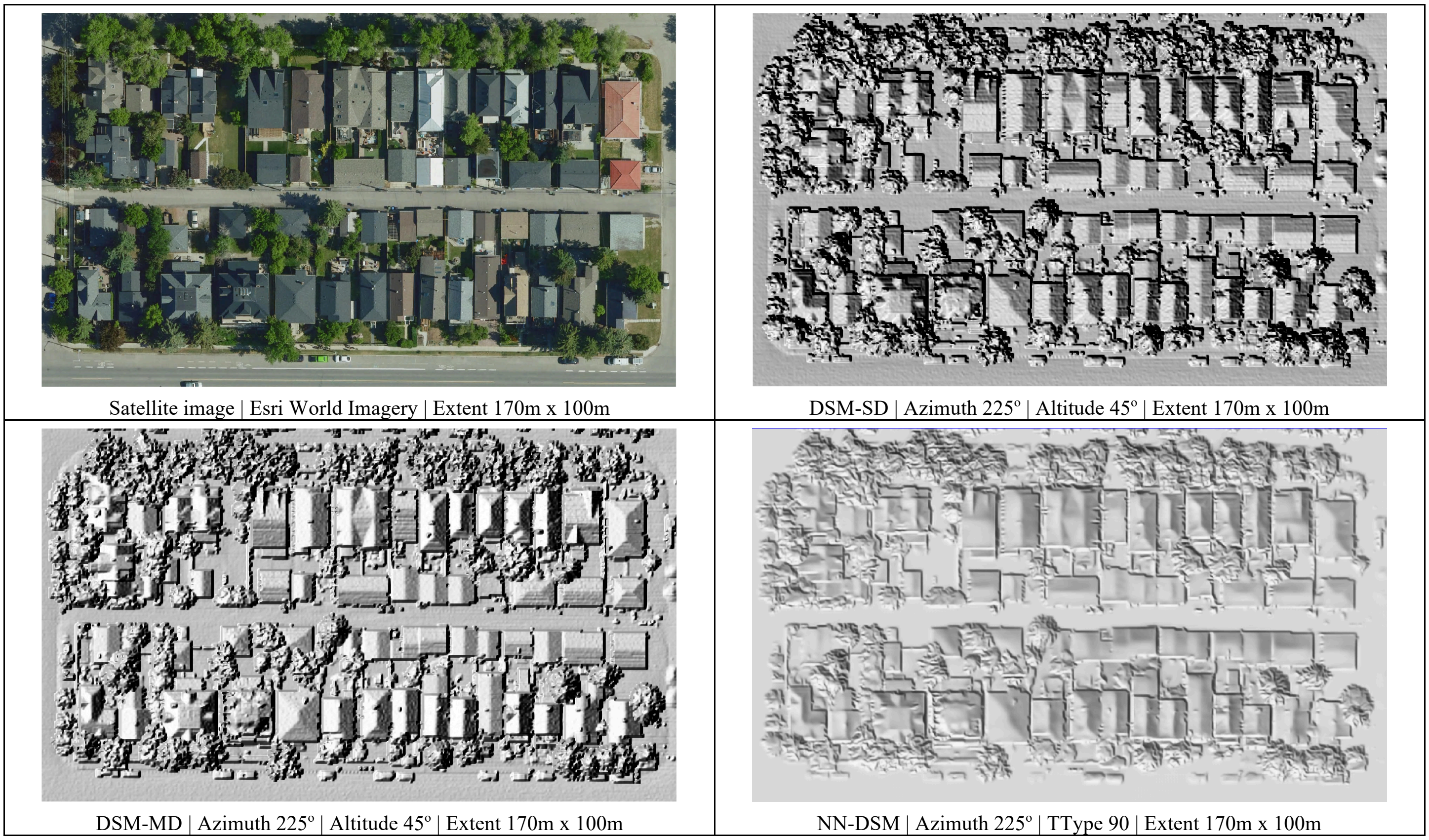

**Figure 8.** Excerpts of Satellite image; Single-direction hillshade; Multi-direction hillshade; and Neural hillshade on the DSM for two residential blocks North of 8 Ave NE and East of 4 Street NE at scale 1:2,000.

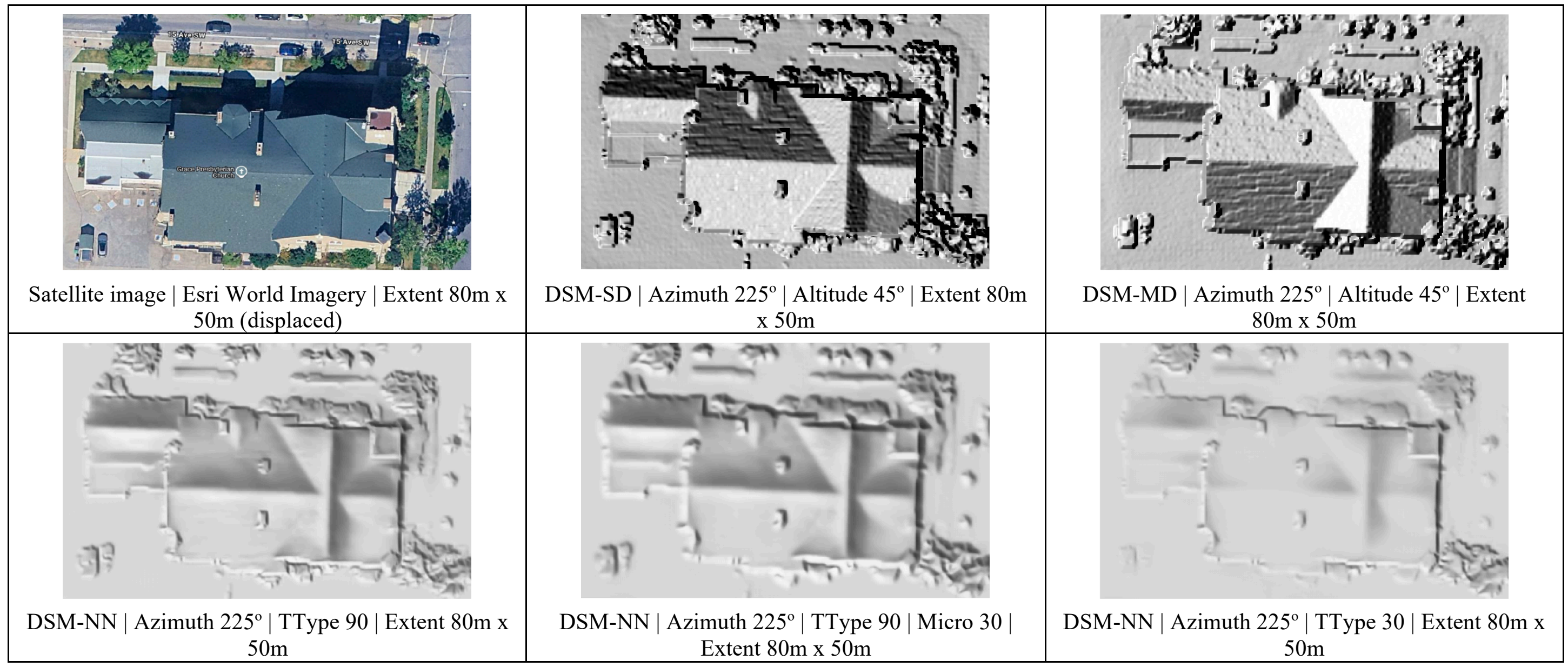

**Figure 9.** Excerpts of Satellite image; Single-direction hillshade; Multi-direction hillshade; and Neural hillshades with various parameter values on the DSM for the Grace Presbyterian Church on the 9 Street SW at scale 1:1,600.

### 4.4 Shading special roof types

Figure 10 presents excerpts from the satellite imagery and hillshades of the Calgary Tower on 9 Ave SW at a scale of 1:1,000. The tower is approximately 190m high and is best described as a circular, saucer-like observation deck with a slightly rounded roof. The analytical hillshades effectively capture the rounded roof form, while the multi-directional hillshade provides a better-balanced representation of the shadows, as expected. By contrast, the neural-based hillshades fail to adequately capture the curved roof and some of the rooftop fixtures, such as the circular skylights, roof hatches, or access/maintenance covers arranged around the white roof of the observation-pod structure. A medium terrain-type value of 60, with or without micro-generalization, produces the clearest hillshade, allowing the roof perimeter and the shape of the ground-level ledges to be distinguished.

| | | |
|---|---|---|
| Satellite image \| Esri World Imagery \| Extent 50m (displaced) | DSM-SD \| Azimuth 225° \| Altitude 45° \| Extent 50m | DSM-MD \| Azimuth 225° \| Altitude 45° \| Extent 50m |
| DSM-NN \| Azimuth 225° \| TType 60 \| Extent 50m | DSM-NN \| Azimuth 225° \| TType 60 \| Micro 30 \| Extent 50m | DSM-NN \| Azimuth 225° \| TType 30 \| Micro 30 \| Extent 50m |

**Figure 10.** Excerpts of Satellite image; Single-direction hillshade; Multi-direction hillshade; and Neural hillshades with various parameter values on the DSM for the Calgary Tower on the 9 Ave SW at scale 1:1,000.

The neural-based hillshades are more successful in capturing the dome of the former Centennial Planetarium. Figure 11 presents excerpts from the satellite imagery and hillshades of the Contemporary Calgary building complex on 11 Street SW at a scale of 1:1,500. Both analytical hillshades capture the dome and the surrounding structures, but they represent the structure near the right edge of the excerpts differently due to differences in shadow rendering. Although the neural-based hillshades capture the dome, they tend to model the surrounding buildings as a surface of smooth terrain patterns. A high terrain type value of 90 without generalization produces the clearest delineation of the surrounding structures.

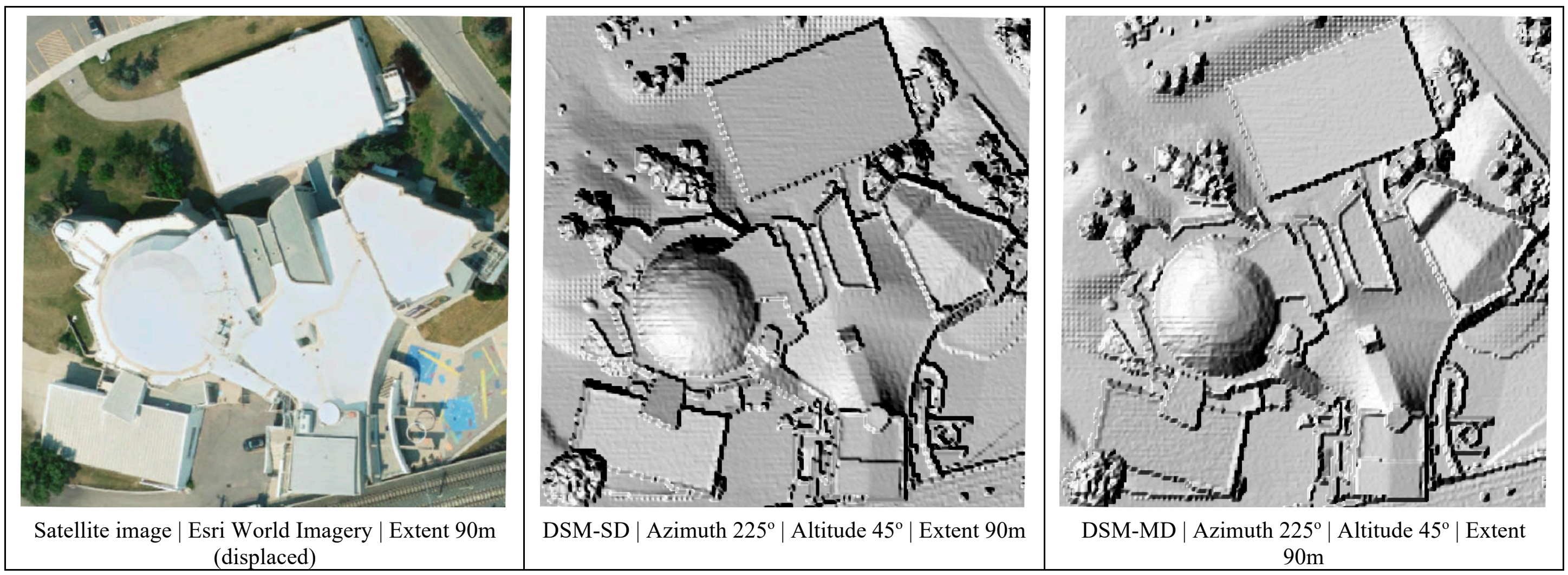

| | | |
|---|---|---|
| Satellite image \| Esri World Imagery \| Extent 90m (displaced) | DSM-SD \| Azimuth 225° \| Altitude 45° \| Extent 90m | DSM-MD \| Azimuth 225° \| Altitude 45° \| Extent 90m |

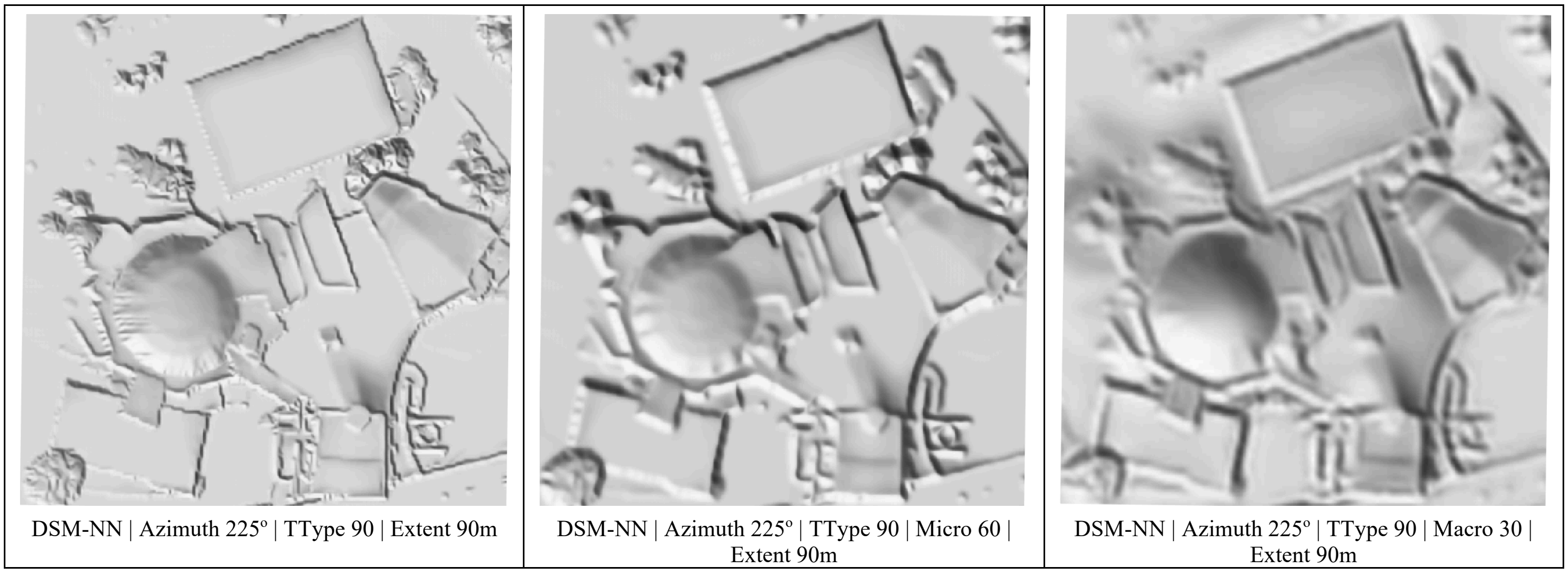

| DSM-NN \| Azimuth 225° \| TType 90 \| Extent 90m | DSM-NN \| Azimuth 225° \| TType 90 \| Micro 60 \| Extent 90m | DSM-NN \| Azimuth 225° \| TType 90 \| Macro 30 \| Extent 90m |
|---|---|---|

**Figure 11.** Excerpts of Satellite image; Single-direction hillshade; Multi-direction hillshade; and Neural hillshades with various parameter values on the DSM for the Contemporary Calgary building complex including the dome (former Centennial Planetarium) on the 11 Street SW at scale 1:1,500.

| Satellite image \| Esri World Imagery \| Extent 30m (displaced) | DSM-SD \| Azimuth 225° \| Altitude 45° \| Extent 30m | DSM-MD \| Azimuth 225° \| Altitude 45° \| Extent 30m |
|---|---|---|
| DSM-NN \| Azimuth 225° \| TType 90 \| Extent 30m | DSM-NN \| Azimuth 225° \| TType 90 \| Micro 30 \| Extent 30m | DSM-NN \| Azimuth 225° \| TType 30 \| Extent 30m |

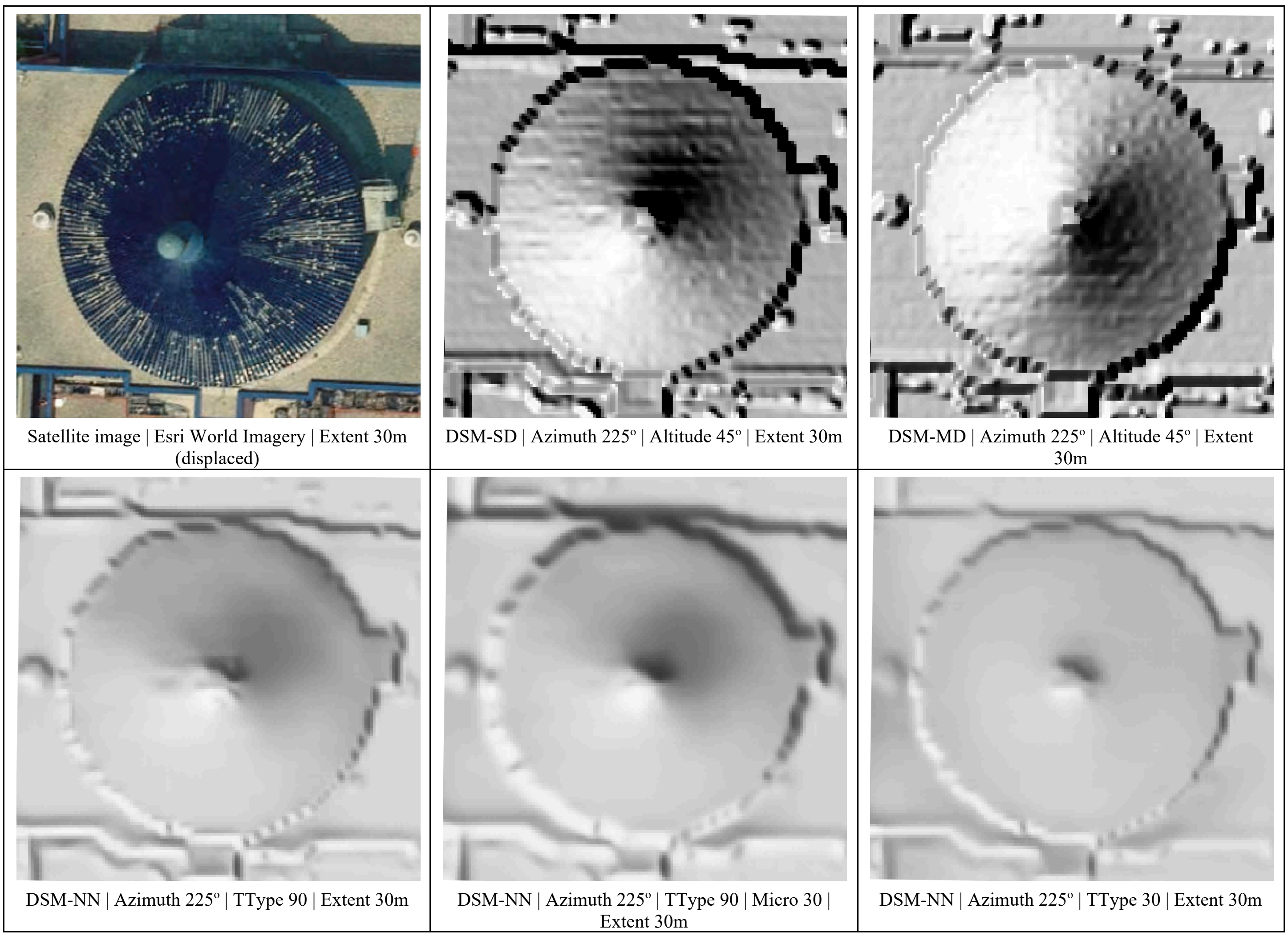

**Figure 12. Excerpts** of Satellite image; Single-direction hillshade; Multi-direction hillshade; and Neural hillshades with various parameter values on the DSM for the conical structure (Cuan Jian roof) of Chinese Calgary Cultural Centre on the 1 Street SW at scale 1:500.

Figure 12 presents excerpts from the satellite imagery and hillshades of the conical roof of the Chinese Calgary Cultural Centre on 1 Street SW at a scale of 1:500. The analytical hillshades effectively capture the conical form, but they depict the roof's base circumference in a cluttered manner. The neural-based hillshades are able to distinguish the conical form at high terrain-type values, particularly at 90, with or without generalization. Reducing the terrain-type value negatively affects the representation of the conical form, although it produces a sharper depiction of the roof's base circumference.

Figure 13 presents excerpts from the satellite imagery and hillshades of the twin towers with pyramidal roof structures at 6 Ave SW and 10 Street SW, at a scale of 1:2,000. The analytical hillshades effectively capture the pyramidal roof forms, as well as the surrounding ground-level structures, streets, and trees. Given the morphological similarity between pyramidal roofs and alpine terrain forms, the neural-based hillshades also capture the roof shapes effectively at high and medium terrain-type values of 90 and 60, with or without generalization. Higher terrain-type values also provide a clearer representation of the surrounding ground-level structures and trees.

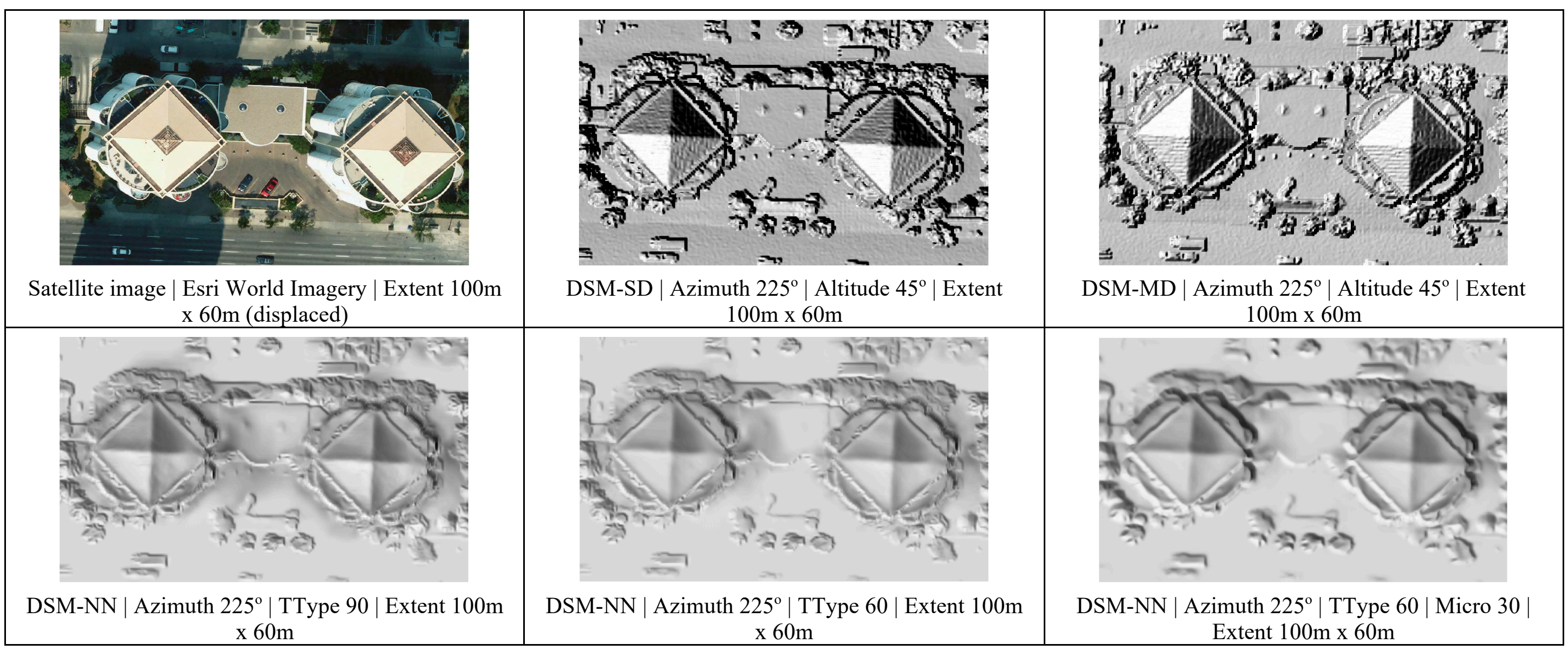

**Figure 13.** Excerpts of Satellite image; Single-direction hillshade; Multi-direction hillshade; and Neural hillshades with various parameter values on the DSM for the twin towers with the pyramid roof structure on the 6 Ave SW and 10 Street SW at scale 1:2,000.

Figure 14 presents excerpts of the satellite imagery and hillshades of the curved glass atrium roof of the CORE Shopping Centre in Calgary, located along 8 Ave SW, at a scale of 1:3,000. The atrium roof is approximately 200m long and 26m wide, forming a prominent elongated and curvilinear structure within the urban block. Both analytical hillshades and the neural-based hillshade generated with a high terrain-type value of 90 effectively depict the overall geometry of the curved roof, including its longitudinal form and distinction from the surrounding built environment. In addition, these hillshades provide a clear representation of adjacent urban features, such as neighbouring buildings, street corridors, and other ground-level structures.

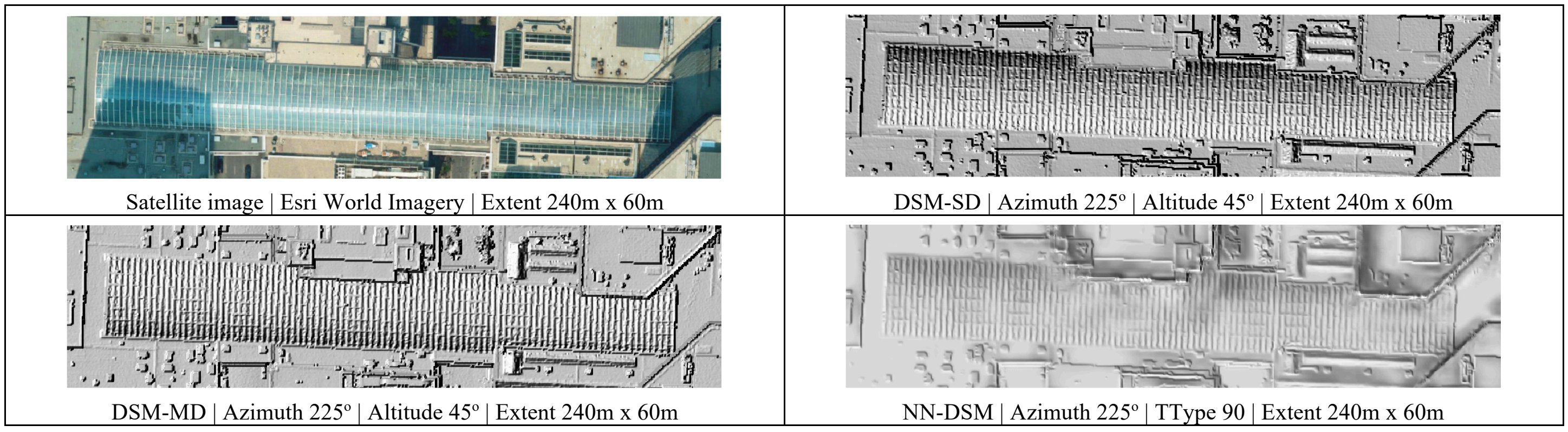

**Figure 14.** Excerpts of Satellite image; Single-direction hillshade; Multi-direction hillshade; and Neural hillshade on the DSM for the curved glass atrium roof at the CORE shopping centre in Calgary on the 8 Ave SW at scale 1:3,000.

### 4.5 Shading vertical forms of high-rise buildings

Figure 15 presents an excerpt of satellite imagery and hillshade visualizations of the 236m tall Bow Tower, located at 6 Avenue SE and Centre Street in Calgary, at a scale of 1:3,000. The analytical hillshades effectively capture structural and morphological details on the roof, as well as ground-level infrastructure, including the curved monument, a large white, wire-mesh head sculpture named 'Wonderland,' located in front of the building. In contrast, the neural-based hillshades provide a more coherent depiction of the tower and surrounding infrastructure at higher terrain-type values. The building outline stands out more clearly than in the analytical hillshade outputs, and vegetation is represented more naturally. However, the curved surface of the monument is not fully captured by the neural-based approach.

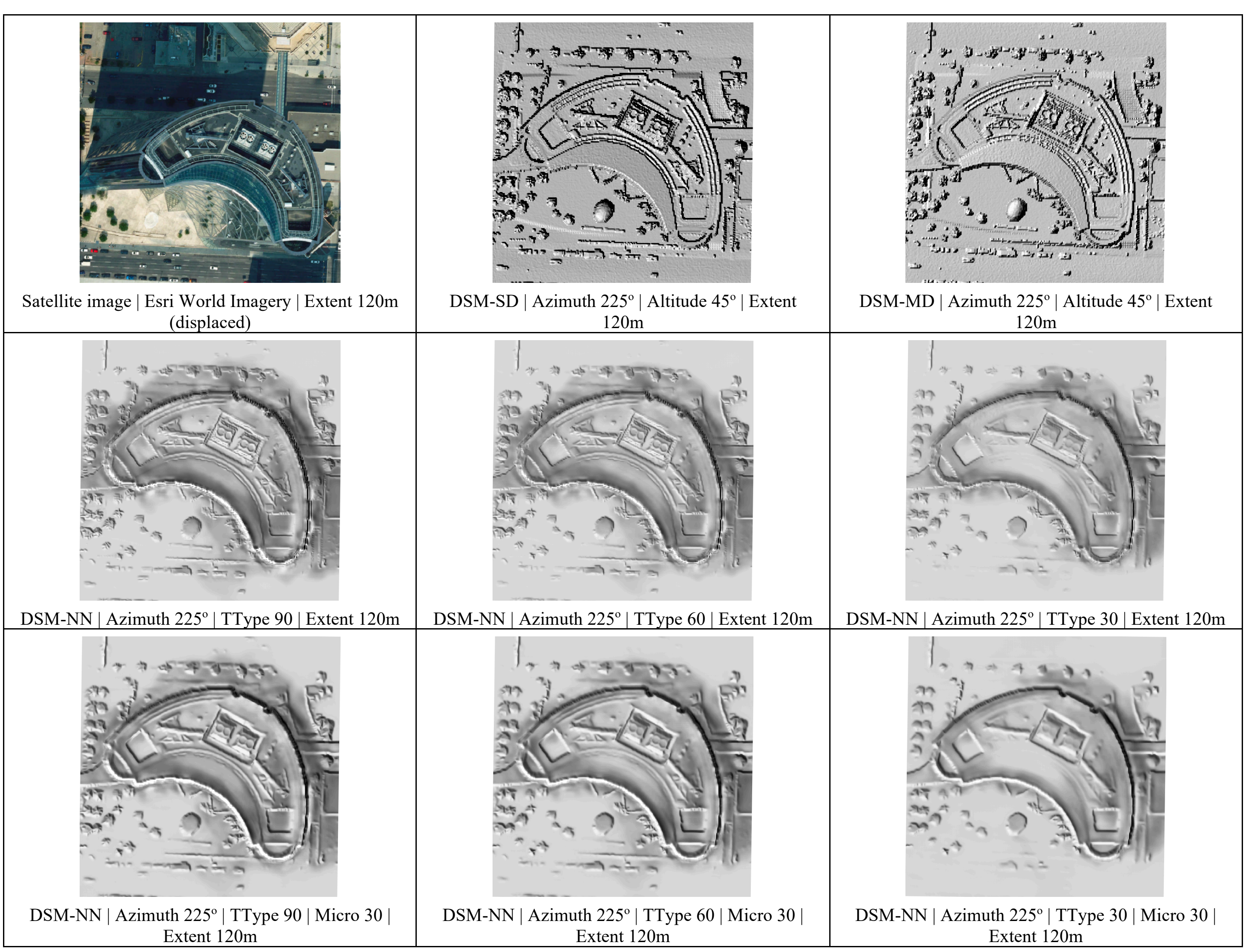

**Figure 15.** Excerpts of Satellite image; Single-direction hillshade; Multi-direction hillshade; and Neural hillshades with various parameter values on the DSM for the Bow Tower on the 6 Ave SE and Centre Street at scale 1:3,000.

Figure 16 presents excerpts of satellite imagery and hillshades for the building blocks between 5 Ave SW and 7 Ave SW, extending west from the Bow Tower to 2 Street SW, at a scale of 1:5,000. Both analytical hillshades provide a sharp representation of infrastructure, including building footprints, roof-level structural and mechanical features, street corridors, vehicles, trees, and other ground-level details. However, except for buildings with large-sloped roof surfaces, these hillshades fail to convey the vertical form and massing of high-rise buildings. In contrast, the neural-based shadings more effectively depict urban morphology in relation to height-related variations in built form, although this comes at the expense of roof and ground-level structural detail. In this case, the best balance is achieved by using a high terrain type value of 90 in combination with a low macro-generalization value of 30.

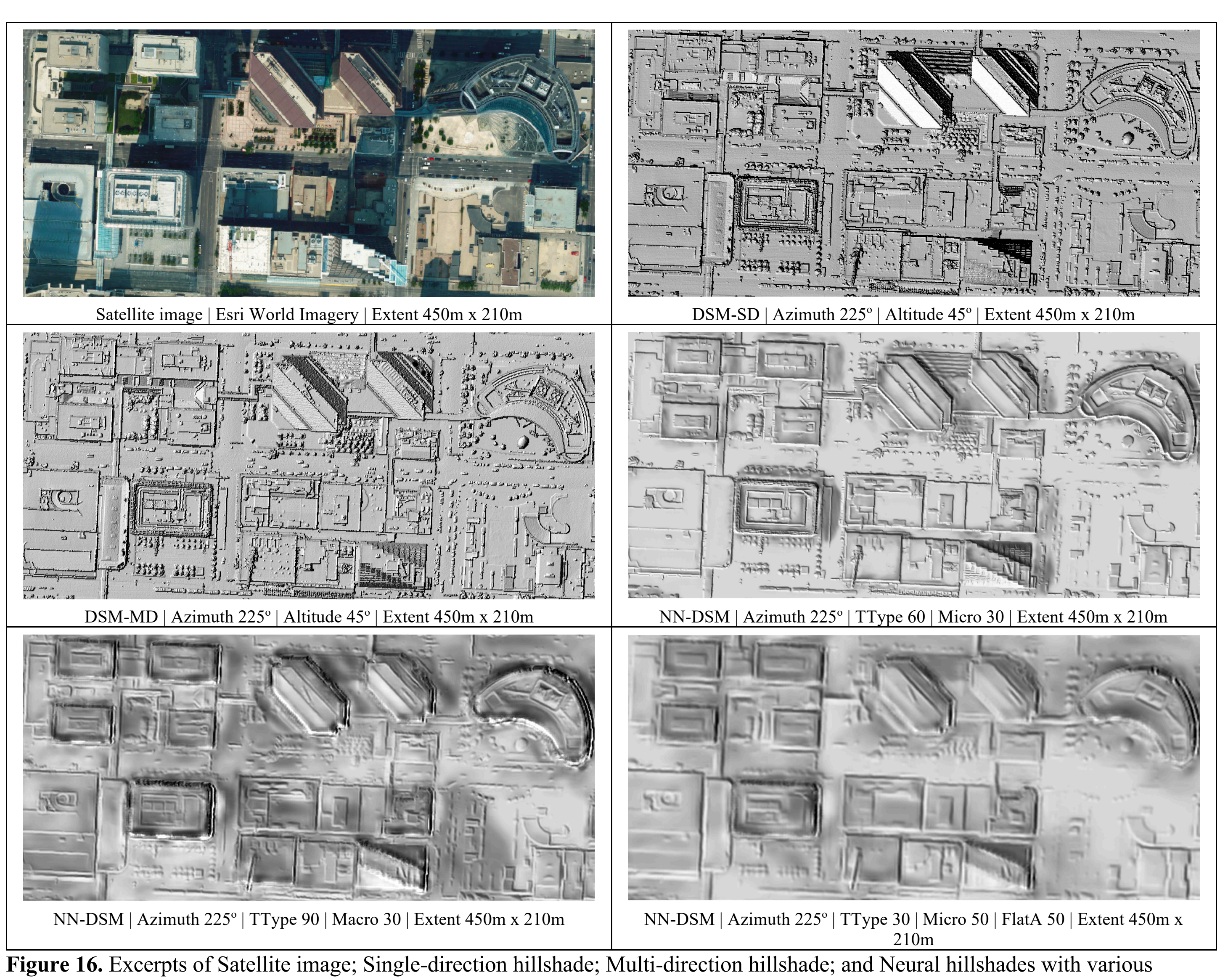

**Figure 16.** Excerpts of Satellite image; Single-direction hillshade; Multi-direction hillshade; and Neural hillshades with various parameter values on the DSM for the building blocks between 5 Ave SW and 7 Ave SW, west of the Bow Tower to 2 Street SW, at scale 1:5,000.

## 5. Discussion

The experiments indicate that neural-based hillshading can depict urban environments, but its performance is highly dependent on the type of urban feature, viewing scale, and parameter configuration. Overall, the neural approach is strongest when urban forms resemble continuous terrain surfaces or when the cartographic objective is a coherent, visually balanced representation rather than precise architectural delineation. High terrain-type values, especially 90, generally produced the most successful results for built environments, particularly for high-rise massing, pyramidal roofs, curved atrium roofs, bridges, and broader urban blocks. In several cases, neural-based shading also represented vegetation more naturally than analytical hillshades, reducing the coarse appearance of trees and producing smoother visual integration between green and grey infrastructure.

At the same time, the results show that neural-based hillshading is not a direct substitute for analytical hillshading where fine urban detail is required. Analytical hillshades more consistently preserved sharp roof ridges, rooftop fixtures, street corridors, pedestrian paths, bridge parapets, riverbank steepness, vehicles, mechanical structures, and other small-scale infrastructure elements. The neural method tended to smooth these features, which improved visual continuity but sometimes reduced cartographic legibility. This limitation was particularly evident in complex architectural forms, such as compound hip roofs, the Calgary Tower observation deck, rooftop fixtures, and small ground-level curved monuments.

The findings also suggest that neural-based hillshading performs differently across urban morphologies. It is effective for generalized vegetation, high-rise building massing, and some special roof types whose shapes resemble natural landforms, such as pyramidal, conical, domed, and elongated curved roofs. However, its effectiveness decreases when the target features depend on crisp edges, fine relief discontinuities, or small constructed details. Parameter tuning can mitigate some of these issues, but it cannot fully overcome the tendency of the neural model to interpret buildings and infrastructure as smoothed terrain-like surfaces. Consequently, neural-based shading, as produced by Eduard, which is not currently trained for relief shading of urban environments, appears most useful as a complementary cartographic technique as it can enhance visual coherence, naturalness, and height perception, while analytical hillshading remains preferable for detailed urban feature extraction and precise morphological interpretation.

## 6. Conclusion

This article proposes an exploratory framework for comparing analytical hillshading and Eduard-based neural relief shading in a high-resolution urban environment. Using 20cm DEM and DSM data for downtown Calgary, the study is designed to show how representation quality depends not only on the shading algorithm itself, but also on the interaction between model type, parameterization, and the built character of the landscape. By keeping the large style scale fixed and varying terrain type, micro and macro generalization, and flat-area detail, the article isolates the conditions under which Eduard can approach or potentially outperform analytical outputs in terms of legibility and cartographic coherence.

The broader conclusion is not that Eduard replaces analytical methods in urban cartography. Rather, the article argues that Eduard remains promising even outside its original alpine training domain, provided that its outputs are interpreted critically and tuned carefully. Its strengths and failures are both informative. Strong outputs suggest that neural relief shading can transfer some cartographic principles across landscape types, while weak outputs demonstrate the need for model training that explicitly includes urban surfaces, building forms, and engineered morphology. Therefore, the study points to the need for future training and evaluation frameworks designed specifically for urban relief shading. To support eventual adoption in urban cartographic practice, Eduard and other neural-based models would need to be further trained on datasets that capture a wide range of urban morphologies, including dense city blocks, detached housing, distinctive roof types, industrial areas, road networks, bridges, and other built features. Such training would help cartographic relief-shading models learn how height, form, edges, and artificial surfaces should be visually generalized in ways that remain cartographically legible, allowing them to serve as tools for producing coherent, natural-looking, and perceptually effective urban terrain representations.

## Acknowledgment

The author gratefully acknowledges the colleagues and students on the evaluation panel for their insightful observations on the countless hillshading excerpts produced for this study.